\documentclass[runningheads]{llncs}
\usepackage{graphicx}
\usepackage{multirow}
\usepackage{epsfig}
\usepackage{amsmath}
\usepackage{amssymb}
\usepackage{algorithm}
\usepackage{algpseudocode}
\usepackage{threeparttable}
\begin{document}

\title{Instance-aware Self-supervised Learning for Nuclei Segmentation}

\author{Xinpeng Xie\inst{1}\thanks{This work was done when Xinpeng Xie was an intern at Tencent Jarvis Lab} \and Jiawei Chen\inst{2}\thanks{Equal contribution} \and Yuexiang Li\inst{2} \and Linlin Shen\inst{1} \and Kai Ma\inst{2} \and Yefeng Zheng\inst{2}}
\authorrunning{X. Xie et al.}
\institute{Computer Vision Institute, Shenzhen University, Shenzhen, China \\
\email{llshen@szu.edu.cn} \and
Tencent Jarvis Lab, Shenzhen, China\\
\email{vicyxli@tencent.com}
}

\maketitle
\begin{abstract}
    Due to the wide existence and large morphological variances of nuclei, accurate nuclei instance segmentation is still one of the most challenging tasks in computational pathology. The annotating of nuclei instances, requiring experienced pathologists to manually draw the contours, is extremely laborious and expensive, which often results in the deficiency of annotated data. The deep learning based segmentation approaches, which highly rely on the quantity of training data, are difficult to fully demonstrate their capacity in this area. In this paper, we propose a novel self-supervised learning framework to deeply exploit the capacity of widely-used convolutional neural networks (CNNs) on the nuclei instance segmentation task. The proposed approach involves two sub-tasks (i.e., scale-wise triplet learning and count ranking), which enable neural networks to implicitly leverage the prior-knowledge of nuclei size and quantity, and accordingly mine the instance-aware feature representations from the raw data. Experimental results on the publicly available MoNuSeg dataset show that the proposed self-supervised learning approach can remarkably boost the segmentation accuracy of nuclei instance---a new state-of-the-art average Aggregated Jaccard Index (AJI) of 70.63\%, is achieved by our self-supervised ResUNet-101. To our best knowledge, this is the first work focusing on the self-supervised learning for instance segmentation.
\end{abstract}

\section{Introduction}
Nuclei instance segmentation provides not only location and density information but also rich morphology features (e.g., magnitude and the cytoplasmic ratio) for the tumor diagnosis and related treatment procedures \cite{ChangH2013}. To this end, many researches have been proposed to establish automated systems for accurate nuclei segmentation. For examples, Xie et al. \cite{XieX2018} utilized Mask R-CNN to directly localize and segment the nuclei instances in histopathological images. Oda et al. \cite{OdaH2018} proposed a deep learning method called Boundary-Enhanced Segmentation Network (BESNet) to segment cell instances from pathological images. The proposed BESNet had similar architecture to U-Net but utilized two decoders to enhance cell boundaries and segment entire cells, respectively. Inspired by \cite{OdaH2018}, Zhou et al. \cite{ZhouY2019} proposed a multi-level information aggregation module to fuse the features extracted by the two decoders of BESNet. The proposed framework, namely Contour-aware Informative Aggregation Network (CIA-Net), achieved an excellent accuracy of nuclei instance segmentation and won the first prize on the Multi-Organ-Nuclei-Segmentation (MoNuSeg) challenge. Although deep learning based approaches achieve outstanding segmentation accuracy for nuclei instances, they share a common challenge for further improvements---the deficiency of annotated data. Due to the wide existence and large morphological variances of nuclei, the annotating of nuclei instances requires experienced physicians to repetitively investigate the histopathological images and carefully draw the contours, which is extremely laborious and expensive. Therefore, the performance of deep learning based approaches suffers from the limited quantity of annotated histopathological images.

Self-supervised learning, as a solution to loose the requirement of manual annotations for neural networks, attracts increasing attentions from the community. The pipeline usually consists of two steps: 1) pre-train the network model on a proxy task with a large unlabeled dataset; 2) fine-tune the pre-trained network for the specific target task with a small set of annotated data. Recent studies have validated the effectiveness of self-supervised learning on multiple tasks of medical image processing such as brain area segmentation \cite{Spitzer_2018}, brain tumor segmentation \cite{Zhuang_2019_MICCAI} and organ segmentation \cite{Zhou_2019_MICCAI}. However, few studies focused on the topic of instance segmentation, which is a totally different area from the semantic segmentation \cite{Jesus2020}. The neural network performing instance segmentation needs to identify not only the object category but also the object instance for each pixel belonging to. Therefore, the self-supervised learning approach is required to implicitly learn to be self-aware of object instance from the raw data for more accurate nuclei segmentation.

In this paper, we propose an instance-aware self-supervised learning approach to loose the requirement of manual annotations in deep convolutional neural networks. The proposed self-supervised proxy task involves two sub-tasks (i.e., scale-wise triplet learning and count ranking), which enforce the neural network to autonomously learn the prior-knowledge of nuclei size and quantity by deeply exploiting the rich information contained in the raw data. The publicly available MoNuSeg dataset is adopted to evaluate the improvements yielded by the proposed proxy task. The experimental results demonstrate that our self-supervised learning approach can significantly boost the accuracy of nuclei instance segmentation---a new state-of-the-art average Aggravated Jaccard Index (AJI) of $70.63\%$ is achieved by the self-supervised ResUNet-101.

\section{Method}
In this section, the proposed instance-aware self-supervised proxy tasks are introduced in details.
\subsection{Image Manipulation}
\begin{figure}[!tb]
    \begin{center}
        \includegraphics[width=0.65\linewidth]{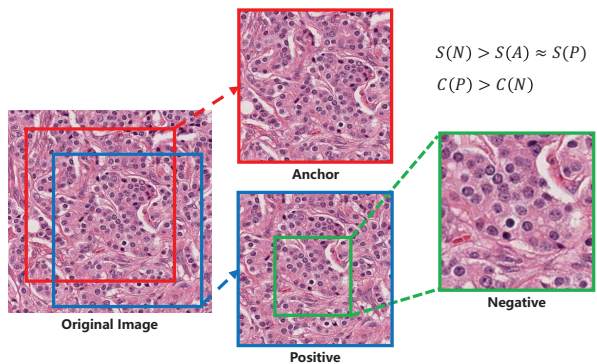}
    \end{center}
    \caption{Image manipulation of a histopathological image. The $S(.)$ and $C(.)$ represent the average nuclei size and the number of nuclei of generated samples (i.e., anchor $A$, positive $P$, and negative $N$), respectively. The two equations reflect heuristic relationship among the samples.}
    \label{triplet_ranking}
\end{figure}

Multiple example learning (e.g., pair-wise and triplet learning), which aims to learn an embedding space that captures dissimilarity among data points, is adopted to encourage neural networks to implicitly learn the characteristics of nuclei instances (i.e., nuclei size and quantity). An example of generating triplet samples for a given histopathological image is shown in Fig.~\ref{triplet_ranking}.

\subsubsection{Nuclei Size.} Specifically, as presented in Fig.~\ref{triplet_ranking}, for a histopathological image of $1000 \times 1000$ pixels from the MoNuSeg dataset, we first crop a patch with $768 \times 768$ pixels (i.e., the red square) as the anchor. Next, we generate a positive sample containing nuclei with similar sizes to the anchor patch by cropping an adjacent patch (i.e., the blue square) with the same size ($768 \times 768$ pixels) from the histopathological image. To better embed the information of nuclei size into self-supervised learning, a negative sample containing nucleus with larger sizes is generated---a sub-patch (i.e., the green square) random cropped from the positive sample and resized to $768 \times 768$ pixels. To increase the diversity of negative samples, the scale of green square is randomly selected from a pool $\{512 \times 512, 256 \times 256, 128 \times 128, 64 \times 64\}$ for each triplet. The anchor, positive and negative samples form a standard triplet data, which is used for the proxy task in self-supervised learning.

\subsubsection{Nuclei Quantity.} The positive and negative samples not only contain nuclei with different sizes ($S$), but also different quantities of nuclei ($C$)---the number of nuclei in negative samples is always lower than that of positive samples. Therefore, we propose to adopt a pair-wise count ranking metric to reflect the difference of nuclei quantity during self-supervised learning.

\subsection{Self-supervised Approach with Triplet Learning and Ranking}
\begin{figure}[!tb]
    \begin{center}
        \includegraphics[width=0.7\linewidth]{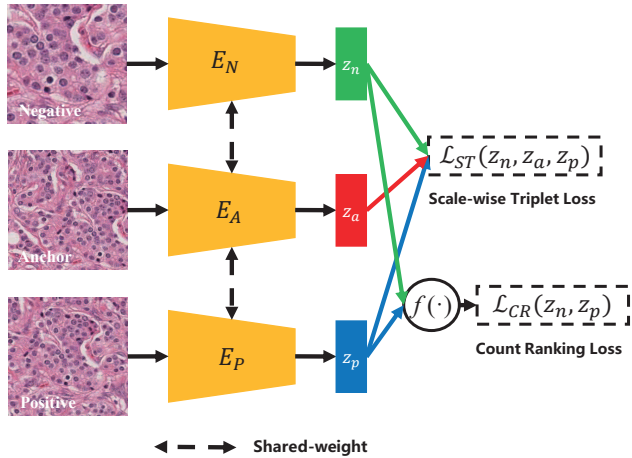}
    \end{center}
    \caption{The pipeline of the proposed self-supervised proxy tasks. The knowledge of nuclei size and quantity is implicitly captured by the scale-wise triplet learning and pair-wise count ranking, respectively.}
    \label{pipeline}
\end{figure}

With the triplet samples (i.e., anchor $A$, positive $P$, and negative $N$), we formulate two self-supervised proxy tasks to pre-train the neural networks for nuclei instance segmentation. The pipeline of the proposed proxy task is illustrated in Fig.~\ref{pipeline}, which consists of three shared-weight encoders supervised by two losses---scale-wise triplet loss and count ranking loss. As aforementioned, the scale-wise triplet learning and count ranking aim to extract features related to knowledge of nuclei size and quantity, respectively. The shared-weight encoders embed the triplet samples into a latent feature space ($Z$), which can be formulated as: $E_A: A \rightarrow z_{a}, \; E_P: P \rightarrow z_{p}, \; E_N: N \rightarrow z_{n}$, where $z_{a}$, $z_{p}$, and $z_{n}$ are $128$-$d$ features.

\subsubsection{Proxy Task 1: Scale-wise Triplet Learning.}
The triplet learning \cite{Schroff2015} encourages samples from the same class to be closer and pushes apart samples from different classes in the embedding space. The proposed approach labels the samples cropped in the same scale with the same class, while treating samples in different scales as different classes. Therefore, the scale-wise triplet loss ($\mathcal{L}_{ST}$) for the embedded triplet features ($z_{a}$, $z_{p}$, $z_{n}$) can be formulated as:
\begin{equation}
    \begin{aligned}
        \mathcal{L}_{ST}\left(z_a,z_p,z_n\right) & =\sum \max \left(0,\ d\left(z_a,z_p\right)-d\left(z_a,z_n\right)+m_1\right)
    \end{aligned}
\end{equation}
where $m_1$ is a margin (which is empirically set to 1.0); $d(.)$ is the squared $L_2$ distance between two features. Regularized by the triplet loss, the network narrows down the perceptional distance between anchor and positive samples in the feature space and enlarge the semantic dissimilarity (i.e., nuclei size) between the anchor and negative samples.

\subsubsection{Proxy Task 2: Count Ranking.}
\begin{figure}[!tb]
    \begin{center}
        \includegraphics[width=\linewidth]{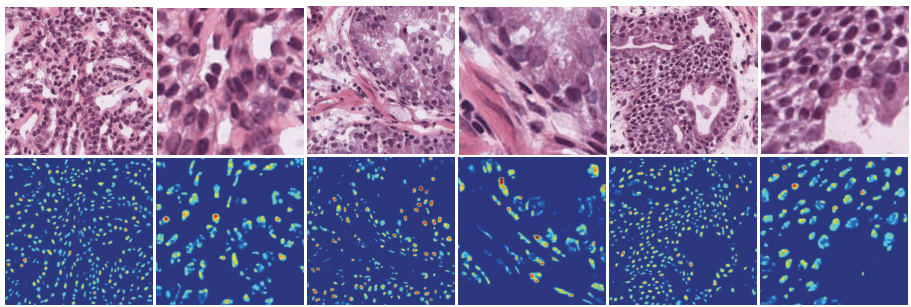}
    \end{center}
    \caption{Density maps of samples containing different quantities of nuclei. The even columns on top represents the negative samples cropped from the positive samples (odd columns). The neural network realizes the variation of nucleus quantity between the positive and negative samples, and activates dense areas in the positive one.}
    \label{map}
\end{figure}

Based on aforementioned observation---the positive sample always contains more nuclei than the negative one, we propose a pair-wise count ranking loss ($\mathcal{L}_{CR}$) to enforce the network to identify the sample containing a larger crowd of nuclei. A mapping function $f$ is first applied to the embedded features ($z_p$, $z_n$) to arrive at a scalar value whose relative rank is known. And in our experiment, $f$ is implemented by a fully convolution layer. Then, the loss function for the embedded features ($z_p$, $z_n$) can be defined as:
\begin{equation}
    \begin{split}\label{CRLoss}
        \mathcal{L}_{CR}&=\sum \max(0, f(z_n) - f(z_p) + m_2)
    \end{split}
\end{equation}
where $m_2$ is a margin (which is empirically set to 1.0). The well-trained network is implicitly regularized to be aware of the nuclei quantity. To further illustrate this, the features extracted from last deconvolution layer are visualized as the nuclei density maps in Fig.~\ref{map}. It can be observed that the density maps of negative samples cropped from positive samples are ordered and sparse, which demonstrate the relative rank created by the mapping function $f$ and the effectiveness of our count ranking loss.

\subsubsection{Objective.}
With the previously defined scale-wise triplet loss $\mathcal{L}_{ST}$ and count ranking loss $\mathcal{L}_{CR}$, the full objective $\mathcal{L}$ for our self-supervised approach is summarized as:
\begin{equation}
    \mathcal{L} = \mathcal{L}_{ST} + \mathcal{L}_{CR}.
\end{equation}

\subsubsection{Fine-tuning on Target Task.}
The recently proposed one-stage framework \cite{CuiY2019} is adopted to perform nuclei instance segmentation. The framework has a U-shape architecture \cite{ronneberger2015u}, which classifies each pixel to three categories (i.e., nuclei body, nuclei boundary and background). We pre-train the encoder of one-stage framework with the proposed proxy tasks to extract instance-aware feature representations and then transfer the pre-trained weights to the target task with a randomly initialized decoder. The widely-used ResNet-101 \cite{He2016} is adopted as the backbone of the encoder. Henceforth, the one-stage framework adopted in this study is referred as ResUNet-101.

\section{Experiments}
The proposed instance-aware self-supervised learning approach is evaluated on the publicly available MoNuSeg dataset to demonstrate its effectiveness on improving segmentation accuracy of nuclei instance.

\begin{table}[!tb]
    \centering
    \caption{AJI (\%) for models finetuned with different amounts of labeled data on the MoNuSeg 2018 test set.}\label{differentamount}
    \begin{tabular}{l|c|c|c|c|c}
        \hline
        \multirow{2}{*}{\bf ResUNet-101} & \multicolumn{5}{c}{AJI(\%)}                                                                             \\\cline{2-6}
                                         & \,\,\,100\%\,\,\,           & \,\,\,70\%\,\,\, & \,\,\,50\%\,\,\, & \,\,\,30\%\,\,\, & \,\,\,10\%\,\,\, \\\hline
        Train-from-scratch               & 65.29                       & 60.33            & 51.45            & 44.32            & 43.58            \\\hline
        ImageNet Pre-trained\,\,\,       & 65.83                       & 62.60            & 53.54            & 48.57            & 48.31            \\\hline
        SSL (Ours)                       & {\bf 70.63}                 & 68.87            & 62.34            & 60.31            & 55.01            \\\hline
    \end{tabular}
\end{table}

\subsection{Datasets}

\subsubsection{MoNuSeg 2018 Dataset \cite{naylor2018segmentation}.} The dataset consists of diverse H$\&$E stained tissue images captured from seven different organs (e.g., breast, liver, kidney, prostate, bladder, colon and stomach), which were collected from 18 institutes. The dataset has a public training set and a public test set. The training set contains 30 histopathological images with hand-annotated nuclei, while the test set consists of 14 images. The resolution of the histopathological images is $1000 \times 1000$ pixels. In our experiments, we separate the public training set to training and validation sets according to the ratio of 80:20. To evaluate the segmentation accuracy of nuclei instance, we adopt the Aggregated Jaccard Index (AJI) \cite{naylor2018segmentation} as the metric. The AJI \cite{naylor2018segmentation} is proved to be a more suitable metric to evaluate the segmentation performance at the object level, which involves matching per ground truth nucleus to one segmented necleus by maximizing the Jaccard index.
\begin{figure}[!htb]
    \begin{center}
        \includegraphics[width=0.95\linewidth]{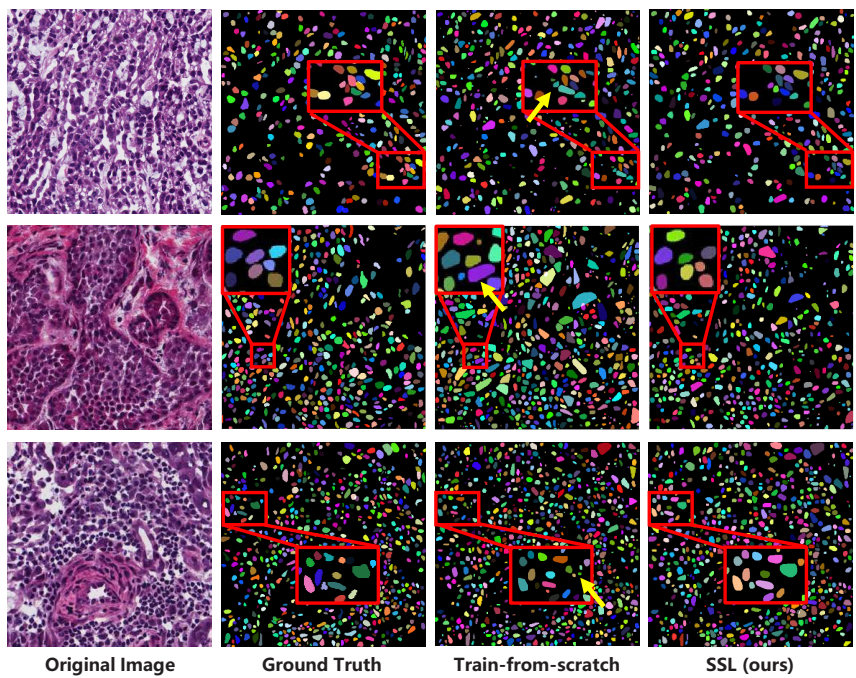}
    \end{center}
    \caption{The nuclei instance segmentation results produced by ResUNet-101 train-from-scratch and self-supervised learning (SSL) pre-trained, respectively.}
    \label{seg_res}
\end{figure}

\subsection{Performance Evaluation}
We first evaluate the improvement yielded by the proposed self-supervised learning proxy tasks to the performance of instance segmentation. The nuclei instance segmentation results of ResUNet-101 trained with different strategies are presented in Fig.~\ref{seg_res}. It can be observed that pre-trained ResUNet-101 produces more plausible segmentation of nuclei instances, especially for the overlapping nuclei marked by yellow arrows, compared to the one trained from scratch. For further evaluation, the AJIs of the two approaches and our self-supervised learning framework finetuned with different amounts of labeled data are evaluated and presented in Table~\ref{differentamount}. Due to the gap between natural and medical images, the ImageNet pre-trained weights yield marginal improvement (e.g., $+0.54\%$ with 100\% annotations) to train-from-scratch. It can be observed that our self-supervised learning proxy tasks significantly and consistently improve the AJI under all conditions, especially with the extremely small quantity (e.g., 10\%) of annotations, i.e., $+11.43\%$ higher than train-from-scratch.

\begin{table}[!tb]
    \centering
    \caption{AJI (\%) for ResUNet-101 trained with different strategies and the top-5 approaches on the MoNuSeg 2018 test set.}\label{competition_results}
    \begin{tabular}{l|c||l|c}
        \hline
        \multicolumn{2}{c||}{\bf Training strategy (ResUNet-101)\,\,\,} & \multicolumn{2}{l} {\bf MoNuSeg 2018 Leaderboard\,\,\,}                                    \\\hline
        Train-from-scratch                                              & 65.29                                                   & Navid Alemi              & 67.79 \\\hline
        ImageNet Pre-trained                                            & 65.83                                                   & Yunzhi                   & 67.88 \\\hline
        Jigsaw Puzzles \cite{noroozi2016unsupervised}                   & 66.68                                                   & Pku.hzq                  & 68.52 \\\hline
        RotNet \cite{gidaris2018unsupervised}                           & 67.61                                                   & BUPT.J.LI                & 68.68 \\\hline
        ColorMe \cite{YLi2019}                                          & 67.94                                                   & CIA-Net \cite{ZhouY2019} & 69.07 \\\hline
        SSL (Ours)                                                      & \multicolumn{3}{c}{\bf 70.63}                                                              \\\hline
    \end{tabular}
\end{table}

For comprehensive quantitative analysis, ResUNet-101 trained with different strategies, including state-of-the-art self-supervised learning approaches \cite{noroozi2016unsupervised,gidaris2018unsupervised,YLi2019}, is evaluated and the results are presented in Table~\ref{competition_results}. The accuracy of the top-5 teams on MoNuSeg 2018 Segmentation Challenge leaderboard\footnote{https://monuseg.grand-challenge.org/Results/} are also involved for comparison. As shown in Table~\ref{competition_results}, the proposed self-supervised learning pre-trained model significantly boosts the accuracy of nuclei instance segmentation ($+5.34\%$), compared with train-from-scratch. We believe that the improvement comes from the prior knowledge of nuclei size and quantity that is learned implicitly in the self-supervised proxy tasks, since our approach outperforms all the listed self-supervised methods, which do not take the instance-related knowledge into consideration. Our self-supervised pre-trained ResUNet-101 is also observed to surpass the winner on the leaderboard (i.e., CIA-Net\cite{ZhouY2019}), which leads to a new state-of-the-art, i.e., $70.63\%$, on the MoNuSeg test set. It is worthwhile to mention that our framework has a much lower computational complexity and fewer network parameters, compared to the CIA-Net, which utilizes DenseNet \cite{Huang_2017_CVPR} as the encoder and has two decoders for nuclei body and boundary, respectively.

\subsubsection{Ablation Study.} To assess the accuracy improvement yielded by each component of our self-supervised learning proxy tasks, we conduct an ablation study. The experimental results are presented in Table~\ref{table_ablation}. Compared to train-from-scratch, fine-tuning from the $\mathcal{L}_{ST}$-only and $\mathcal{L}_{CR}$-only pre-trained weights improves the segmentation accuracy by $+4.35\%$ and $+4.80\%$, respectively. Since jointly pre-training on the two sub-tasks (i.e., $\mathcal{L}_{ST}$ and $\mathcal{L}_{CR}$) increases the diversity of feature representation learned by neural networks, it provides the highest improvement, i.e., $+5.34\%$.
\begin{table}[!tb]
    \centering
    \caption{Performance produced by different self-supervised proxy tasks on the MoNuSeg 2018 test set.}\label{table_ablation}
    \begin{tabular}{c|c|c|c|c}
        \hline
        Setup      &\,\,\, ResUnet \,\,\,& \,\,\,+$\mathcal{L}_{ST}\,\,\,$ & \,\,\,+$\mathcal{L}_{CR}$ \,\,\,& \,\,\,+$\mathcal{L}_{ST}$+$\mathcal{L}_{CR}$ \,\,\,\\\hline
        {AJI (\%)} & 65.29   & 69.64               & 70.09               & {\bf 70.63}                            \\\hline
    \end{tabular}
\end{table}

\subsubsection{Validation on Another Dataset.} The evaluation results on the Computational Precision Medicine (CPM) dataset can be found in Table~\ref{CPM_results}.

\begin{table}[!htb]
    \centering
    \caption{Dice score (\%) and AJI (\%) on the Computational Precision Medicine (CPM) dataset*. A 5-fold cross validation is conducted. Apart from the AJI, we also evaluate the Dice score, which proposed by the CPM competition. An average Dice of 86.36\% is achieved by our self-supervised ResUNet-101, which is comparable to the winner of CPM 2018 competition (i.e., 87.00\%).}\label{CPM_results}
    \begin{tabular}{l|c|c|c|c|c|c}
        \hline
                                   & \,\,\, {Fold 1} \,\,\, & \,\,\, {Fold 2} \,\,\, & \,\,\, {Fold 3} \,\,\, & \,\,\, {Fold 4} \,\,\, & \,\,\, {Fold 5} \,\,\, & \,\,\, {Average} \,\,\, \\\hline\hline
        \multicolumn{7}{l}{\bf Dice score}\\\hline
        Train-from-scratch & 85.29                 & 83.99                 & 84.12                 & 82.89                 & 86.05                 & 84.47                   \\\hline
        SSL (Ours)                           & 86.54                 & 85.18                 & 85.84                 & 86.08                 & 88.14                 & {\bf 86.36}             \\\hline\hline
        \multicolumn{7}{l}{\bf AJI}\\\hline
        Train-from-scratch  & 74.43                 & 72.79                 & 72.80                 & 71.03                 & 75.60                 & 73.33                   \\\hline
        SSL (Ours)                         & 76.34                 & 74.56                 & 75.37                 & 75.84                 & 78.83                 & {\bf 76.19}             \\\hline
    \end{tabular}
    \begin{tablenotes}
    \item * https://wiki.cancerimagingarchive.net/pages/viewpage.action?pageId=37224869
    \end{tablenotes}
\end{table}

\section{Conclusion}
In this paper, we proposed an instance-aware self-supervised learning framework for nuclei segmentation. The proposed proxy consists of two sub-tasks (i.e., scale-wise triplet learning and count ranking), which enable the neural network to implicitly acquire the knowledge of nuclei size and quantity. The proposed self-supervised learning proxy tasks were evaluated on the publicly available MoNuSeg dataset and a new state-of-the-art AJI (i.e., $70.63\%$) was achieved.

\section*{Acknowledge}
This work is supported by the Natural Science Foundation of China (No. 91959108 and 61702339), the Key Area Research and Development Program of Guangdong Province, China (No. 2018B010111001), National Key Research and Development Project (2018YFC2000702) and Science and Technology Program of Shenzhen, China (No. ZDSYS201802021814180).

\bibliographystyle{splncs04}

\end{document}